\documentclass{article}

\usepackage[letterpaper, left=1in, right=1in, top=1in, bottom=1in]{geometry} %

\usepackage[utf8]{inputenc}
\usepackage[T1]{fontenc}

\usepackage{graphicx} %
\usepackage{hyperref}
\usepackage{url}
\usepackage{graphicx}
\usepackage{pdfpages}
\usepackage{caption}
\usepackage{subcaption}
\usepackage{tabularx}
\usepackage{amsmath}

\usepackage[round]{natbib}

\usepackage{amsmath,amsthm,amssymb}

\usepackage{graphicx}
\usepackage{array}
\usepackage{tabularx}
\usepackage{makecell}
\usepackage{subcaption}
\usepackage{xcolor}
\usepackage{tikz}
\usepackage{verbatim}
\usepackage{footmisc}
\newcommand{\cmmnt}[1]{}
\usepackage{enumitem}
\usepackage{xspace} %

\usepackage{natbib}
\setcitestyle{round}

\ifx\lemma\undefined
\newtheorem{theorem}{Theorem} 
\newtheorem{lemma}[theorem]{Lemma}

\else
\fi

\title{ChatGPT Participates in a Computer Science Exam}

\author{
  Sebastian Bordt \\
  Department of Computer Science\\
  University of Tübingen \\
  \texttt{sebastian.bordt@uni-tuebingen.de} \\
  \and
  Ulrike von Luxburg \\
Department of Computer Science and Tübingen AI Center\\
  University of Tübingen \\
  \texttt{ulrike.luxburg@uni-tuebingen.de} \\
}

\begin{document}

\maketitle

\begin{abstract} 
We asked ChatGPT to participate in an undergraduate computer science exam on ''Algorithms and Data Structures''. The program was evaluated on the entire exam as posed to the students. We hand-copied its answers onto an exam sheet, which was subsequently graded in a blind setup alongside those of 200 participating students. We find that ChatGPT narrowly passed the exam, obtaining 20.5 out of 40 points. This impressive performance indicates that ChatGPT can indeed succeed in challenging tasks like university exams. At the same time, the questions in our exam are structurally similar to those of other exams, solved homework problems, and teaching materials that can be found online and might have been part of ChatGPT's training data. Therefore, it would be inadequate to conclude from this experiment that ChatGPT has any understanding of computer science. We also assess the improvements brought by GPT-4. We find that GPT-4 would have obtained about 17\% more exam points than GPT-3.5, reaching the performance of the average student. The transcripts of our conversations with ChatGPT are available at \url{https://github.com/tml-tuebingen/chatgpt-algorithm-exam}, and the entire graded exam is in the appendix of this paper. 
\end{abstract}
    
\section{Introduction}

Many have noted that the capabilities of ChatGPT\footnote{\url{chat.openai.com}}, a novel chat-bot model by OpenAI, are so impressive that the model might even be able to succeed in challenging real-world tasks like university exams \citep{meta23}. Indeed, there is already existing evidence to suggest that this might be the case \citep{bommarito2022gpt,choi2023chatgpt,kung2023performance,frieder2023mathematical}. However, apart from one study by legal scholars \citep{choi2023chatgpt}, existing evaluations on university exams probe the model only on a subset of the task for which it might be particularly suited (for example, excluding all questions that contain images). In addition, evaluation of the model's responses is often not blind, which can be problematic because ChatGPT is known to produce strange answers that are subject to interpretation. As such, despite much discussion about the topic, there is to this point little systematic evidence regarding the capabilities of ChatGPT on university exams \citep{mitchell2023really}. 

{\bf We present the results of a simple but rigorous experiment} that evaluates the capabilities of ChatGPT on an undergraduate computer science exam about algorithms and data structures. We conducted this experiment alongside the regular university exam, which allowed us to evaluate the model's responses in a blind setup jointly with those of the students. We posed the different exam questions in a simple standardized format that allowed ChatGPT to give clear answers to all exam questions. We then wrote the answers from ChatGPT by hand onto an exam paper and put it into the stack of student papers. Teaching assistants then graded the answers given by ChatGPT without knowing of this experiment.

{\bf The result of the experiment is that ChatGPT would have narrowly passed the exam, obtaining 20.5 out of 40 points.} This is impressive but also highlights the limitations of the current version of the model. In particular, the performance of the model was worse than the performance of the average student who participated in the exam (the average student obtained about 24 points, compare Table~\ref{tab:points}). The mixed performance of ChatGPT is interesting insofar as the exam is relatively standardized. Similar exams are being written all around the world, and lots of information about the topics covered in the exam is available on the internet. 

{\bf We also assess the improvements brought by GPT-4.} We find that ChatGPT with the GPT-4 base model would have obtained about 17\% more points on the exam than ChatGPT with the GPT-3.5 base model, reaching the performance of the average student. The observed improvement of 17\% puts our computer science exam roughly in the 50th percentile of the different exams considered in the GPT-4 technical report (compare Table 1 in \citet{openai2023gpt4}).

\section{Related Work}
\label{sec:related_work}

\begin{table}
\centering
\Large
\begin{tabular}{l|c|c|c|c|c|c|c|c|c|c|c}
ChatGPT-3.5 & \multicolumn{11}{c}{\includegraphics[width=0.73\textwidth]{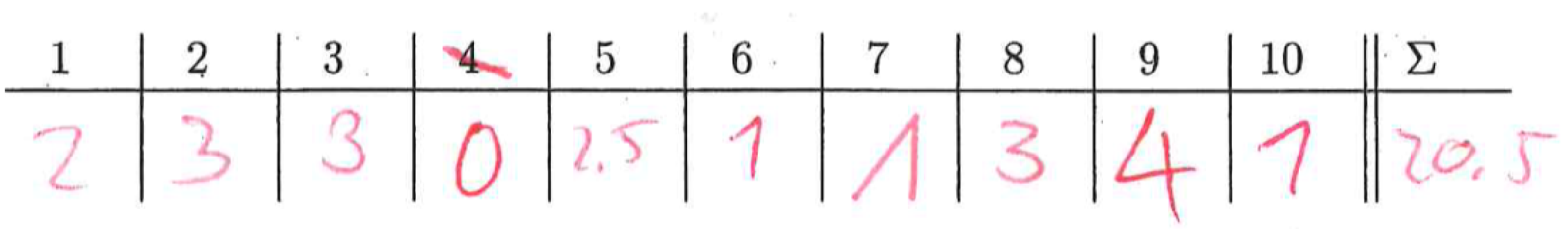}}\vspace{-0.25em} \\
\hline\vspace{-0.75em}\\
ChatGPT-4 & \multicolumn{11}{c}{\hspace{-0.75em}\includegraphics[width=0.715\textwidth]{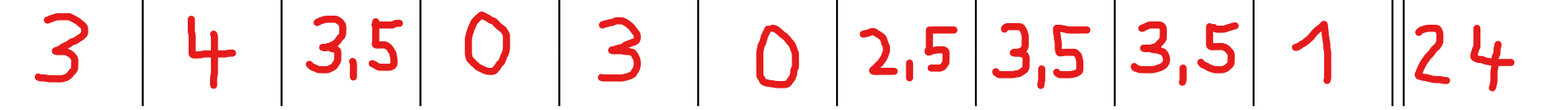}} \\[1pt]
\hline
Student Mean & $\,$ 2.3 & 2.9  &  3.1 & 2.2 & 2.9 & 2.9 & 2.6 & 2.9 & 1.7 & 1.1 & 23.9 \\
\end{tabular}
\caption{The number of points obtained by ChatGPT on the 10 different exercises in our exam, in comparison with the average number of points obtained by the 200 students who participated in our exam. The first row depicts the points obtained by ChatGPT with the GPT-3.5 base model. This is the result of the main experiment discussed in this paper, where model responses were blindly graded alongside student responses. %
ChatGPT-3.5 obtained 20.5 out of a maximum of 40 points. The second row depicts the points obtained by ChatGPT with the GPT-4 base model. Here, the model responses were graded according to the same grading scheme used in the main experiment, but grading was not blind. We gauge that ChatGPT-4 would have obtained about 24 points, reaching the performance of the average student. The third row depicts the average number of points obtained by the 200 students who participated in our exam.}
\label{tab:points}
\end{table}

Large language models, now often called foundation models, have shown remarkable success across a variety of tasks \citep{brown2020language,ouyang2022training}. In particular, they have shown diverse problem-solving capabilities, for example, on university-level math problems \citep{drori2022neural,frieder2023mathematical}. At the same time, even very large language models suffer from certain well-documented limitations that also plagued earlier iterations \citep{marcus2020gpt,bender2021dangers,borji2023categorical}. A large variety of different works now use foundation models to solve specific tasks, such as coding \citep{noever2020chess,chen2021evaluating,madaan2023learning}. Similarly, an emergent literature aims to enhance the  capabilities of language models, for example by giving them access to a numerical calculator \citep{schick2023toolformer}. Because of the dramatic increase in the capabilities of language models in recent years, there is currently a need to develop novel benchmarks to assess these models \citep{srivastava2022beyond,binz2023using,shiffrin2023probing}. For a more comprehensive review of the literature, we refer the reader to \citep{kalyan2021ammus,srivastava2022beyond,meta23}.

\section{Experimental Design}
\label{sec:experimental_design}

\subsection{The Exam}

We consider an exam of an introductory class on algorithms and data structures that Bachelor of Science students in Computer Science typically take in their second year (the third semester in the German university system). The exam covers topics such as sorting algorithms, graph traversal, and dynamic programming. Overall, the topics covered in the exam are being taught in a similar way all across the world. The exam is contained in a single latex text file, and even the figures in the exam are produced by latex code that consists of text only. This is important for our study, since it allows us to pass the figures as text to ChatGTP and evaluate the language model on the entire exam. The exam contains different types of questions, including multiple-choice questions, writing small proofs, writing pseudo-code, and drawing graphs. The entire exam, complete with the answers of ChatGPT and grading remarks, can be found in Appendix \ref{apx:exam}. The idea to conduct this experiment came only after designing the exam, and we did not bias the exercises in the exam towards the capabilities of ChatGPT.

\subsection{Posing the Exam Questions to ChatGPT}
\label{sec:posing_questions}

\begin{figure}
     \centering
     \begin{subfigure}[b]{0.4\textwidth}
         \centering
         \includegraphics[width=\textwidth]{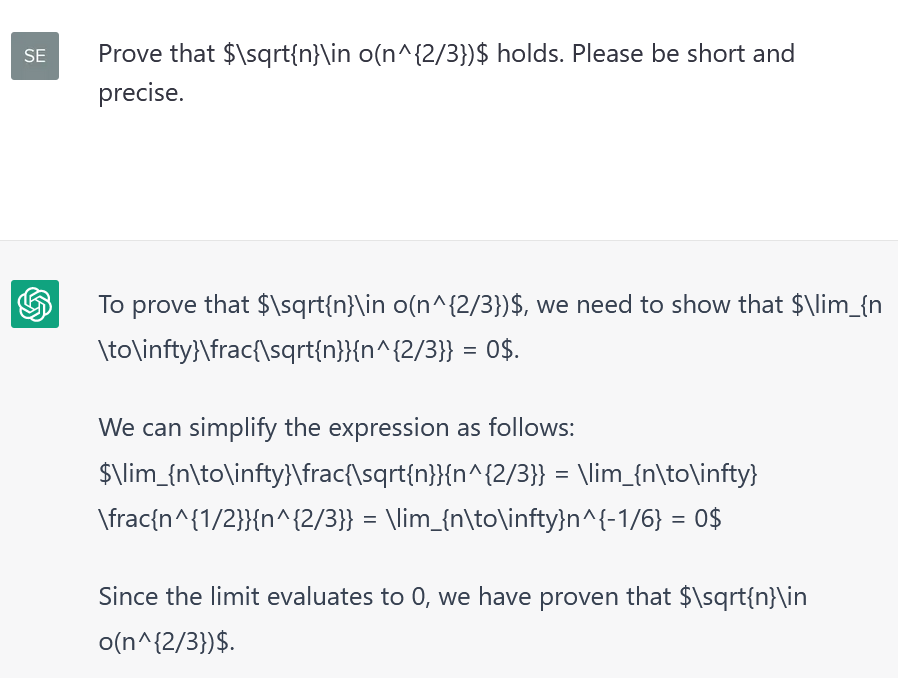}
         \caption{Our conversation with ChatGPT.}
     \end{subfigure}
     \begin{subfigure}[b]{0.5\textwidth}
         \centering
         \includegraphics[width=\textwidth]{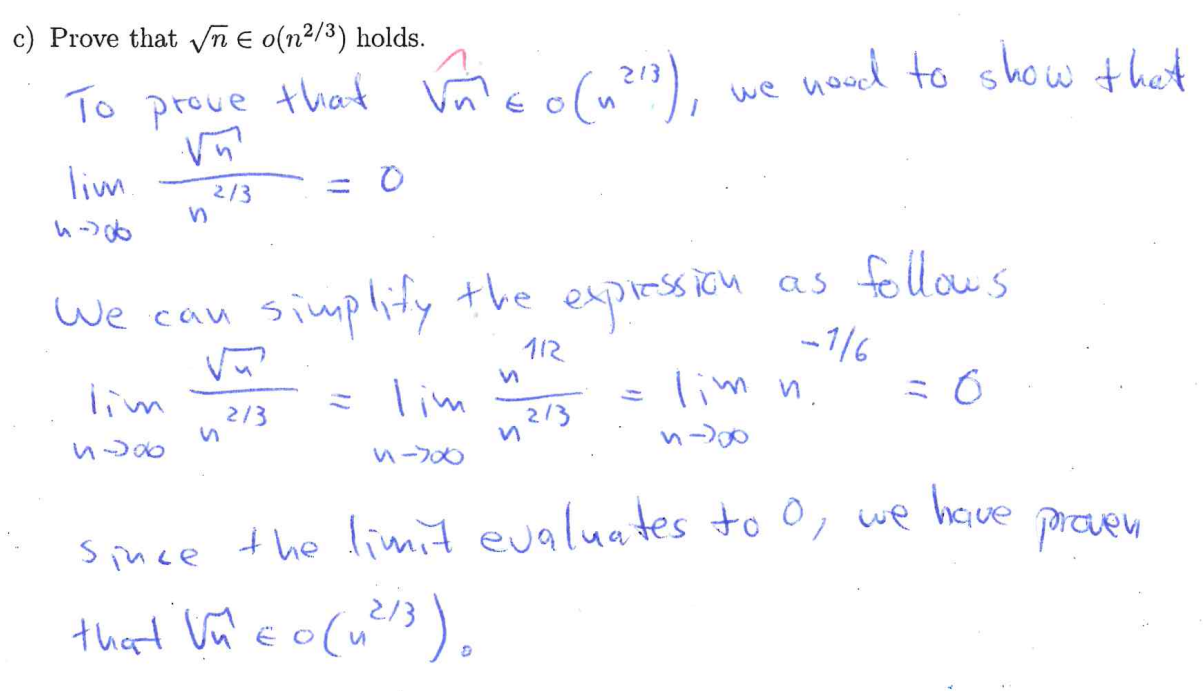}
         \caption{The corresponding part of the written exam.}
     \end{subfigure}
        \caption{We obtained the responses of ChatGPT in different conversations with the model. We then hand-copied the responses on an exam sheet, so that they could be blindly graded with the student exams. In the depicted example, the model provides a proof of the fact that $\sqrt{n}\in o(n^{2/3})$.}
        \label{fig:example}
\end{figure}

{\center\includegraphics[width=0.9\textwidth]{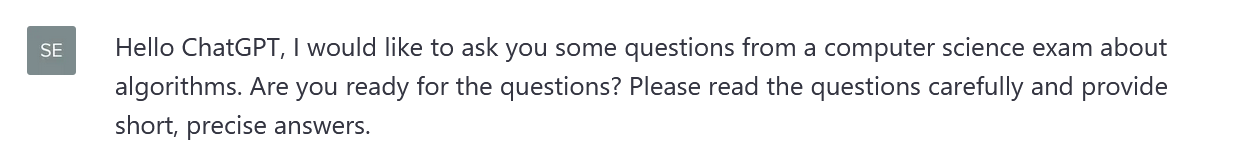}$\,$\\}

We posed the exam questions in 19 different conversations with the model, relying on the latex source file of the exam. All questions were posed to the default model on 16.2.2023. We were able to elicit clear answers to all exam questions within a couple of hours. Because it is known that the prompt can significantly impact the output of ChatGTP (''prompt engineering''), we decided to pose the exam as simply as possible. We did not perform any chain of thought reasoning \citep{wei2022chain}, since similar techniques might also enhance student performance. We informed the model that we were asking questions from a computer science exam about algorithms and asked it to provide short, precise answers. During the entire process, we did not attempt to engineer the prompts in order to steer the model toward better or worse answers. Our only objective was that the model  would give clear answers to all questions (for multiple-choice questions, this would mean that the model clearly stated which option it wanted to choose). The transcript of our conversations with the model is available at \url{https://github.com/tml-tuebingen/chatgpt-algorithm-exam}.

Some of the exam questions involved maths, pseudo-code, or graphs. In this case, we simply prompted the model with the latex source code from the exam, as in the following example:

{$\quad$\includegraphics[width=0.45\textwidth]{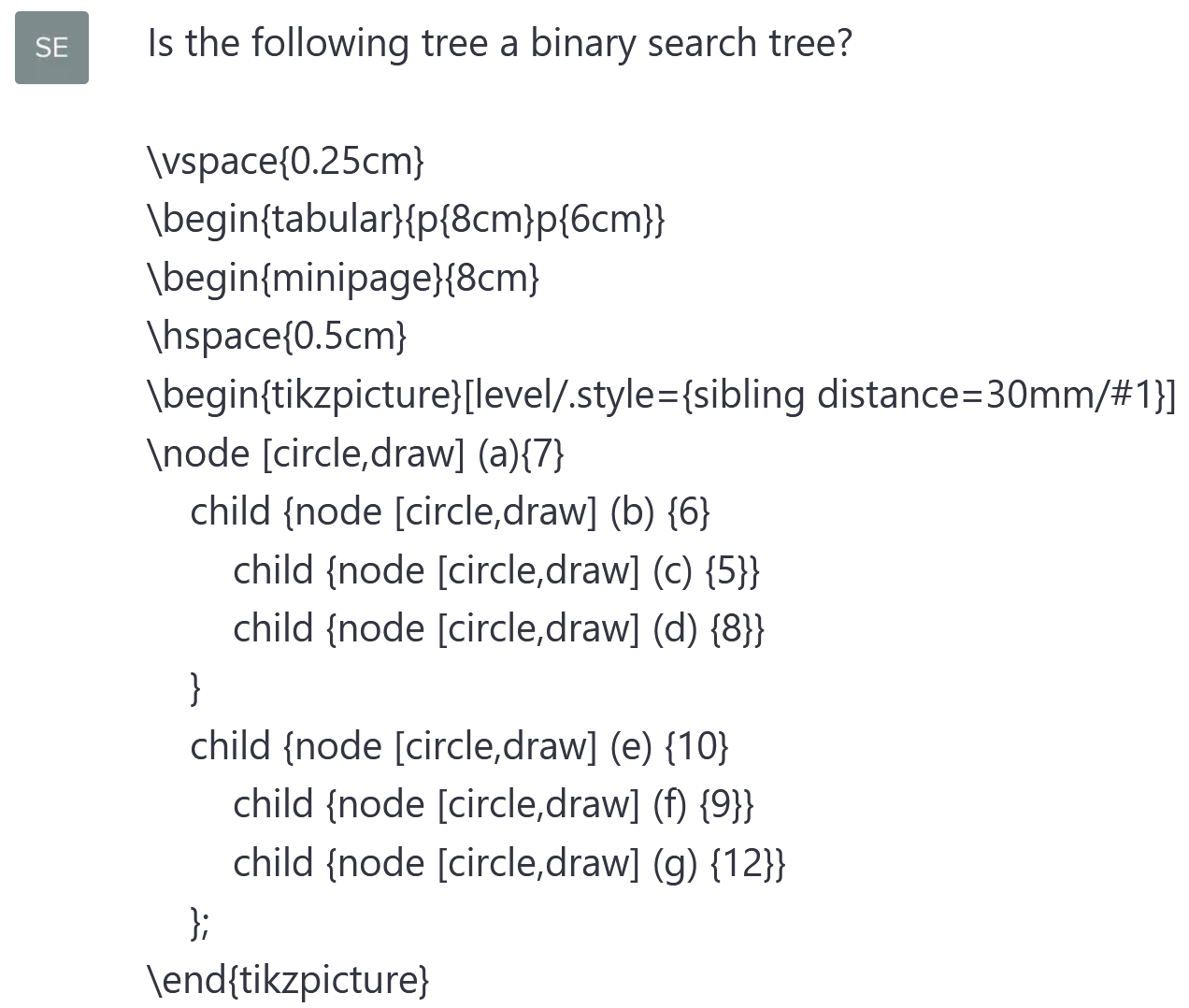}$\,$\\}

Remarkably, the model gave clear and well-defined answers to all of our questions.\footnote{In response to exercise 4b), the model drew some relatively strange figures in order to indicate different graphs. We simply copied these figures onto the exam paper so that a teaching assistant could decide whether they were worth any points (they weren't).}  When we asked the model to write a small proof, for example, it responded with latex equations. This is depicted in Figure~\ref{fig:example}. Similarly, when we asked the model to complete pseudo-code, it completed the given pseudo-code in a valid way. In two questions, the model had to draw a graph. Both times, the model responded with a valid {\tt tkiz}-graph that could be rendered in latex (this is depicted in Figure \ref{fig:graph_out}).

After conducting the conversations with the model, we copied the responses by hand onto an exam paper sheet (the students wrote the exam on these exact paper sheets, too). In doing so, we did of course ''render'' all latex output of the model onto the paper sheet. A concrete example where the model responded with a {\tt tkiz}-graph is depicted in depicted Figure \ref{fig:graph_out}.

\subsection{Grading}

The exam was taken by about 200 students, who all wrote their answers in hand on the exam sheets. We simply slipped the sheet with ChatGPT's responses among all the other student's sheets. The exam was jointly graded on one day by 10 teaching assistants, who all worked in the same room at the same time. Each of the 10 teaching assistants graded one of the 10 exercises on the exam. The study was blind, meaning that the teaching assistants did not know that one of the solutions had been produced by ChatGTP. An experimenter was present when the teaching assistants graded the exams, and there was no evidence that anyone noticed the exam by ChatGPT. Indeed, while the responses of ChatGPT were sometimes strange, some selected student responses were (unfortunately) not much more meaningful either. 
The graded exam can be found in Appendix \ref{apx:exam}.

\section{Results}
\label{sec:results}

\begin{figure}
     \centering
     \begin{subfigure}[b]{0.35\textwidth}
         \centering
         \includegraphics[width=\textwidth]{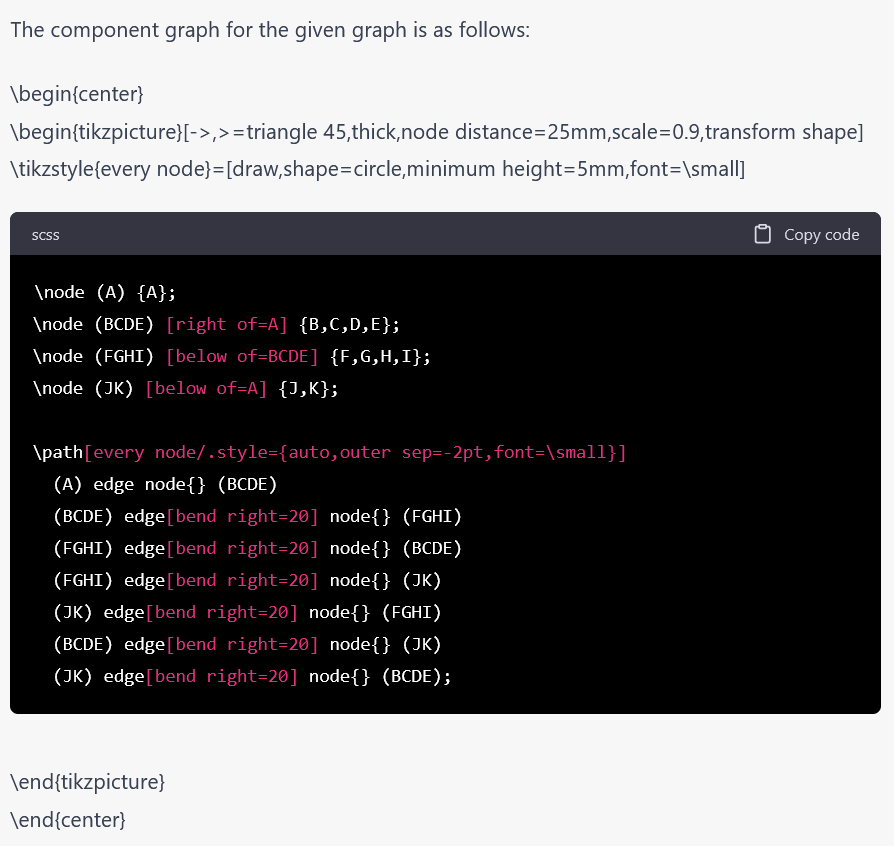}
         \caption{Our conversation with ChatGPT.}
         \label{fig:graph_chatgpt}
     \end{subfigure}
     \begin{subfigure}[b]{0.5\textwidth}
         \centering
         \includegraphics[width=\textwidth]{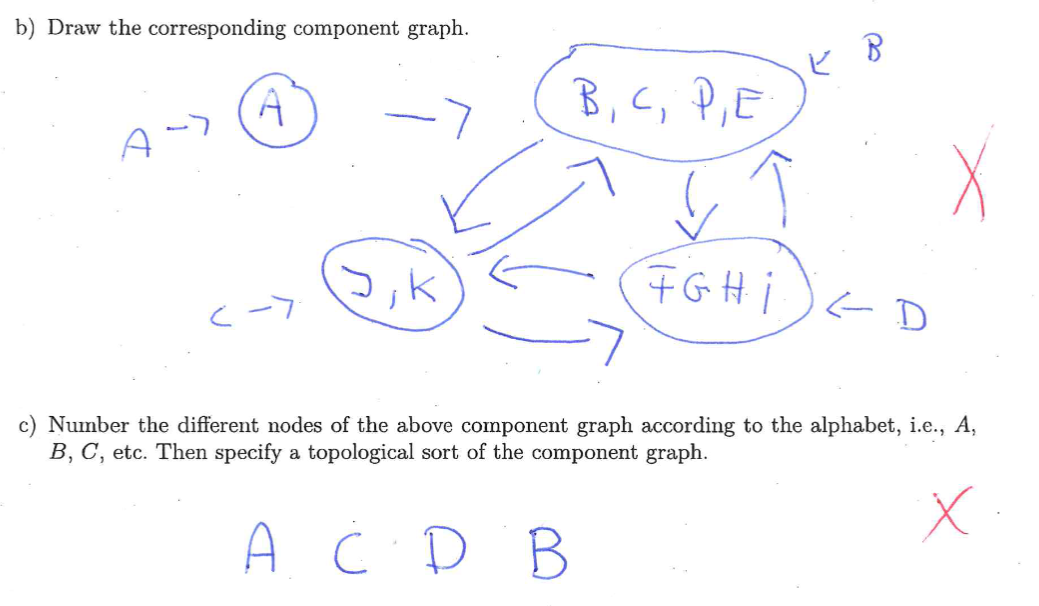}
         \caption{The corresponding part of the written exam.}
         \label{fig:graph_exam}
     \end{subfigure}
        \caption{ChatGPT successfully produced structured output, including pseudo-code and {\tt tkiz}-graphs. In the depicted example, ChatGPT responds with a {\tt tkiz}-graph that is valid in the sense that it corresponds to a valid latex command that can be rendered into a graph. We then copied the graph by hand onto the exam sheet. The response by ChatGPT is not the correct answer to the exam question.}
        \label{fig:graph_out}
\end{figure}

In this section, we discuss the results of the main experiment with GPT-3.5. The improvements brought by GPT-4 are discussed in the next section. The main result is that ChatGPT obtained 20.5 out of 40 points and passed the exam. Since a minimum of 20 points was required in order to pass the exam, ChatGPT passed only very narrowly\footnote{In exercise 3a), the teaching assistant did not recognize a partial mistake made by ChatGPT. Had the mistake been recognized, ChatGPT would have gotten 20 points.} 

On some parts of the exam, the performance of ChatGPT was rather impressive. One example is the small proof depicted in Figure \ref{fig:example}. Here, the model shows that $\sqrt{n}\in o(n^{2/3})$. While this is not very difficult to show, the model's response is flawless and even gives the impression that the model ''understands'' the statement and how to prove it. In addition to this example, ChatGPT showed strong performance on all exercises where the solution involved pseudo-code (exercises 3d, 8f, and 9). In particular, the model provided a very good answer to exercise number 9 on dynamic programming. This is in contrast to our students, who tended to struggle with this exercise. For a comparison of the points obtained by ChatGPT versus the average number of points obtained by the students who participated in the exam, see  Table \ref{tab:points}. 

On other parts of the exam, ChatGPT produced wrong and sometimes strange answers (consider the  gibberish character mix in exercise 4b in the Appendix). In particular, the model struggled on all exercises where that involved structured output that was not pseudo-code (exercises 4a, 5a, 6b, 7a, and 7b). These were often exercises that asked the students to explicate the workings of standard algorithms. For the students, these were among the simplest exercises in our exam. ChatGPT, however, struggled with them. One example of this is depicted in Figure \ref{fig:graph_out}. In this exercise, the solution, by definition, is known to be a directed acyclic graph. ChatGPT, however, produces a graph with lots of loops. In general, it stands out that the model hardly managed to solve any of the questions that involved drawing graphs, even if it always produced valid \texttt{tikz}-pictures.

The performance of the model across the different multiple-choice exercises in our exam was comparable to its overall performance (after correcting for random guessing).

\section{GPT-4}

In this section, in addition to the main experiment considered in this paper, we assess the improvements brought by GPT-4. The GPT-4 technical report compares GPT-4 and GPT-3.5 on a number of different exams and reports large performance gains \citep{openai2023gpt4}. However, because the data sets used in the report are not available, it is hard to replicate and assess these results. It has, however, been noted that there is evidence for testing on the training data \citep{snakeoilexams2023}.

We posed the exam questions to ChatGPT with the GPT-4 base model ('ChatGPT-4') in the same way as we posed them in the main experiment (described in Section \ref{sec:posing_questions}). We again copied the answers on an exam sheet, available with the transcripts at \url{https://github.com/tml-tuebingen/chatgpt-algorithm-exam}. The only difference with the main experiment is that grading was not blind - we might interpret the result as an estimate of the points that would have been obtained under blind grading. 

In total, ChatGPT-4 obtained 24 out of 40 points. This is an improvement of 3.5 points, or 17\%, over ChatGPT with the GPT-3.5 base model. Interestingly, this means that ChatGPT-4 is on par with the performance of the average student in our exam. While the improvement might seem small, it does in fact mean that ChatGPT-4 was able to answer some of the more challenging multiple-choice questions that the previous version of the model struggled with. ChatGPT-4 was also able to correctly derive the shortest paths on a weighted graph given to it as a {\tt tkiz}-graph (exercise 7b). As such, our simple experiment indicates that GPT-4 does indeed constitute a notable improvement over GPT-3.5.

\section{Discussion}
\label{sec:discussion}

At this point, there is already a large debate about the capabilities and risks of large language models \citep{webson2021prompt,bommasani2021opportunities}. One part of this debate is about the properties of current systems \citep{acoup2023}. The other part is about the question of whether some of the demonstrably existing limitations of the current systems will or will not be overcome in future iterations \citep{marcus2020gpt,bender2021dangers}. With our simple experiment, we intend to provide a single data point of relatively high quality for the study of the currently existing systems.

The results of our experiment are in line with existing research that documents both the impressive capabilities of large language models, as well as their severe limitations \citep{binz2023using,borji2023categorical}. We would like to highlight that the fact that ChatGPT was able to pass our exam does not imply that it has any understanding of computer science, in the way in which we might expect it from a human who was able to pass the exam \citep{mitchell2023really}. Many of the topics covered in the exam are extensively documented on the internet. For example, they all have relatively high-quality Wikipedia entries, and many teaching materials including examples of solved homework problems and assignments can be found online. As such, it is reasonable to assume that ChatGPT has seen many exercises and solutions that are similar to those in our exam during training. In general, much more research is required in order to understand the abilities and limitations of models like ChatGPT \citep{binz2023using,shiffrin2023probing}.

\section*{Acknowledgements}

This work has been supported by the German Research Foundation through the Cluster of Excellence “Machine Learning – New Perspectives for Science" (EXC 2064/1 number 390727645), the BMBF Tübingen AI Center (FKZ: 01IS18039A) and the International Max Planck Research School for Intelligent Systems (IMPRS-IS).

\bibliography{literature}

\clearpage
\appendix

\section{The Exam}
\label{apx:exam}

\begin{figure}


\section*{Supplementary material}

# A   The Exam

# 1st Exam "Algorithms and Data Structures"

February 16, 2023

Name: Chatarina Gopelte

Matriculation Number: 3812071

I hereby declare that I will only use the one-sided handwritten A4 cheat sheet written by me. I feel physically able to take the exam. If I am not registered for the exam, my participation is subject to change. I will not take any copies of the exam out of the room (neither printed nor handwritten).

Signature: Chatarina Gopelte

Score achieved:

| 1 | 2 | 3 | 4 | 5 | 6 | 7 | 8 | 9 | 10 | Σ |
|---|---|---|---|---|---|---|---|---|---|---|
| 2 | 3 | 3 | 0 | 2.5 | 1 | 1 | 3 | 4 | 1 | 20.5 ✓ |

## Important Information:

- Please place your ID card and student ID card on the table.
- Turn off and pack away your cell phones completely.
- **You only need to answer 9 of the 10 questions.** If you answer all 10 questions, we will only consider the best 9 of them in the final sum.

  Therefore, you should focus on exactly 9 questions, and only attempt the remaining question if you have time.
- Check that your exam paper includes all 15 pages.
- Write your name on each sheet.
- If you need more paper, we will provide it to you. Do not use your own paper.
- Clearly indicate on new sheets which task you are working on.
- Hand in your cheat sheet along with the exam.
- Do not write with a pencil.
- The total processing time is 120 minutes.

Good luck!

NAME: _____

**Aufgabe 1: Landau Notation and Growth of Functions** (2.5 + 0.5 + 1 Punkte)

a) Decide for each of the following function pairs $f$ and $g$, in which asymptotic relationship they stand to each other. Cross out EVERY box that represents a true statement. You receive 0.5 points for every completely correctly filled row in the table.

| $f$ | $g$ | $f \in o(g)$ | $f \in \mathcal{O}(g)$ | $f \in \Theta(g)$ | $f \in \Omega(g)$ | $f \in \omega(g)$ | |
|---|---|---|---|---|---|---|---|
| $sin(n)$ | $n$ | ☒ | ☒ | ☐ | ☒ ☐ | ☐ | 0.5 no cross! |
| $n^2 + cos(n)$ | $n^2 + \frac{2}{n}$ | ☒ | ☒ | ☐ | ☒ | ☐ | |
| $(\sqrt{5})^n$ | $4^n$ | ☒ | ☐ | ☐ | ☒ | ☐ | |
| $log(n^n)$ | $n^{2/3}$ | ☐ | ☐ | ☒ | ☒ | ☐ | 0 |
| $n \bmod 20$ | $n^2$ | ☒ | ☒ | ☐ | ☐ | ☐ | 0.5 |

b) Which of the following statements implies that $f(n) \in \Omega(g(n))$ holds? Cross exactly one box.

☒ $f(n) \leq 4g(n)$ for all $n \geq 1$

☐ $f(n) \geq 4g(n)$ for all $n \geq 136$

☐ $\lim_{n \to \infty} \frac{f(n)}{g(n)} = 0$

☐ None of the above statements.

c) Prove that $\sqrt{n} \in o(n^{2/3})$ holds.

To prove that $\sqrt{n} \in o(n^{2/3})$, we need to show that

$$\lim_{n \to \infty} \frac{\sqrt{n}}{n^{2/3}} = 0$$

We can simplify the expression as follows

$$\lim_{n \to \infty} \frac{\sqrt{n}}{n^{2/3}} = \lim_{n \to \infty} \frac{n^{1/2}}{n^{2/3}} = \lim_{n \to \infty} n^{-1/6} = 0$$

Since the limit evaluates to 0, we have proven that $\sqrt{n} \in o(n^{2/3})$.

NAME: _____

**Aufgabe 2: Multiple Choice - Diverse Topics**  (Score per sub-task: 0.5 Punkte)

For all questions, exactly one answer choice is correct and therefore only one box is to be filled in. If you fill in more than one box for a question, your answer for that question will be considered incorrect. For each correctly answered question, you will receive 0.5 points, for an incorrectly answered question 0 points.

a) If we could find the pivot element in the QuickSort algorithm in $O(1)$ time, then we would get the best guarantee on the running time with

☐ the arithmetic mean     ☒ the median   ✓

b) The connected components of an undirected graph are uniquely determined

☒ Correct     ☐ Incorrect   ✓

c) The principle of "Memoization" in dynamic programming means

☐ we store the dynamic-programming recurrence in a static array so that it may be applied at each level of recursion.

☒ we save the solutions of smaller subproblems if we need them again.

☐ the dynamic-programming algorithm can remember the current location in the recursion tree.

☐ we will remember the recursions occurring in dynamic programming for future exams.  ✓

d) Which of the following properties is *not* a desired property of a hash function $h(x)$?

☐ $h(x)$ should be computable in $O(1)$ time.

☐ The values of $h(x)$ should be within the size of the hash table.

☐ For different $x$, $h(x)$ should also assume different values.

☒ If $x_1$ to $x_n$ are the elements to be inserted into the hash table, then $h(x_1)$ to $h(x_n)$ should be evenly distributed over the natural numbers.   ✓

e) In the Relax-operation of the Floyd-Warshall algorithm, if we replace the minimum with the maximum, then the algorithm will not find the shortest, but the longest paths between pairs of vertices.

☒ Correct     ☐ Incorrect   ✗

f) If you increase every edge weight by the same positive constant, then the solution to the single-source shortest-path problem will not change.

☒ Correct     ☐ Incorrect   ✗

g) A connected, unweighted Graph $G = (V, E)$ has a unique minimum spanning tree.

☐ Correct     ☒ Incorrect   ✓

h) If $T(n) = 2T(n/3) + \sqrt{n}$ holds, then we also have

☐ $T(n) = \Theta(\sqrt{n})$

☐ $T(n) = \Theta(n^2)$

☒ $T(n) = \Theta(n^{\log_3(2)})$

☐ $T(n) = \Theta(n^{\log(3)})$   ✓

NAME: _____

**Aufgabe 3: Data structures** (1 + 1 + 0.5 + 1.5 Punkte)

a) Perform the DecreaseKey(0, 3) operation on the following Max-Heap (in this operation, the key of the root element is set to 3). Draw the resulting Max-Heap.

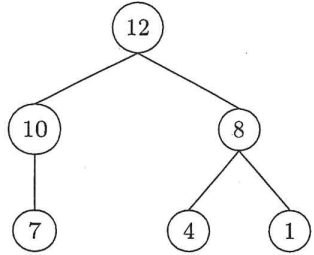

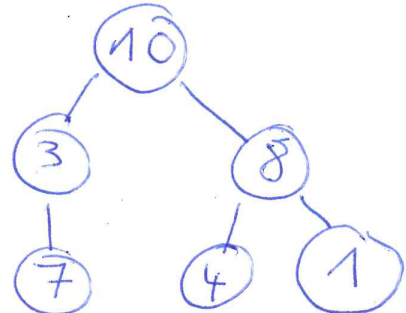

b) Give an instance of a knapsack problem with 4 different items and weights where the greedy algorithm does not find the optimal solution. The greedy algorithm selects in each step the item with the best value-weight ratio that still fits in the knapsack.

| Item | Weight | Value |
|------|--------|-------|
| 1 | 3 | 8 |
| 2 | 2 | 7 |
| 3 | 4 | 9 |
| 4 | 5 | 10 |

Knapsack capacity: 7

c) Is the following tree a binary search tree?

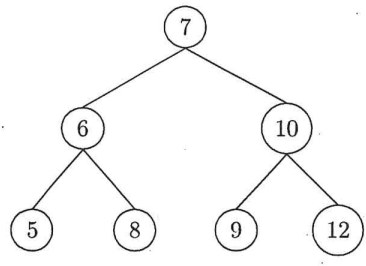

☐ Yes
☒ No

d) Let $T$ be a binary search tree with root node $x$. It is assumed that access to the children of any node can be obtained with $x.left$ and $x.right$. It holds that $x.left == NIL$ or $x.right == NIL$ if no left or right subtree exists. Provide the pseudo-code for an algorithm that outputs the nodes of the binary search tree in sorted order from largest to smallest using $PRINT$.

**function** PRINTBSTREVERSESORTED(Root $x$)

    if $x \ != NIL$ then
        PRINTBSTREVERSESORTED(x.right)
        PRINT(x.key)
        PRINTBSTREVERSESORTED(x.left)
    END IF

**end function**

NAME:

**Aufgabe 4: Hashing and MST**  (1 + 3 Punkte)

a) Consider hashing with linear probing and the hash function $h(x) = (7 \cdot x + 2) \mod 11$. The corresponding hash table consists of an array of length 11. Insert the elements 2, 3, 7, 15 and 1 in this order into the hash table.

| 15 | 7 | 2 | 3 | 1 |  |  |  |  |  |  |
|----|---|---|---|---|--|--|--|--|--|--|

✓p

b) Consider the following algorithm. The algorithm receives as input a connected, undirected, weighted graph $G = (V, E, c)$ with edge weight function $c : E \to \mathbb{R}_{\geq 0}$.

**function** ALTERNATIVEMST
 $E_T = \emptyset$
 Sei $s$ beliebiger Knoten in $G$
 **for** $v \in V \setminus \{s\}$ **do**
  Let $e = (v, w)$ be the edge of $v$ with minimal weight $c(e)$ that does not belong to $E_T$
  $E_T \leftarrow E_T \cup \{e\}$
 **end for**
 **return** $(V, E_T)$
**end function**

Is AlternativeMST correct, that is, does AlternativeMST actually calculate a Minimum Spanning Tree of $G$? Prove or disprove that AlternativeMST calculates a MST.

AlternativeMST is not correct and does not always calculate an MST. Here is a counterexample

$G = $

```
    1 -3- 2
   /      |
  1       1
   \      |
    3 -1- 2
```

Running AlternativeMST on $G$ with $s=1$ would produce

```
1 -- 3 -- 2
          |
          1
          |
3 ------- 2
```

2.

However, the actual MST would be

```
1 -- 3 -- 2
|         |
1         1
|         |
3 -- 1 -- 2
```

NAME: _____

**Aufgabe 5: DFS** (1 + 1.5 + 0.5 + 0.5 + 0.5 Punkte)

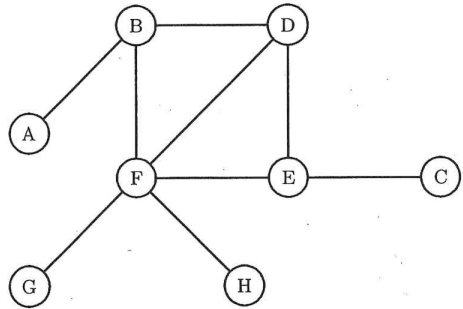

a) Perform the DFS algorithm on the above graph, starting at node $A$. Assume that the DFS algorithm visits nodes with smaller letters first when there is a choice. Give the order in which the various nodes are visited.

A B D F E C G H ~~F~~

b) Now give the discovery time and the finishing time of the different nodes. Start with the discovery time $t = 1$ for node $A$.

| Knoten $v$ | Discovery-Zeit($v$) | Finishing-Zeit($v$) |
|---|---|---|
| A | 1 | 16 |
| B | 2 | 9 |
| C | 14 | 15 |
| D | 3 | 8 |
| E | 10 | 13 |
| F | 4 | 7 |
| G | 11 | 12 |
| H | 5 | 6 |

ff.

1 / 1.5

c) An undirected and unweighted graph is given as an adjacency matrix. Then the running time of the DFS algorithm is $O(|V| + |E|)$.

☐ Correct ☒ Incorrect ✓

d) Let $G = (V, E)$ be a directed graph. Which of the following algorithmic problems can be solved with DFS? Put exactly one cross.

☐ Finding a source node.
☐ Finding a sink node. ✓
☒ Both problems.
☐ Neither of the two problems.

e) Let $G = (V, E)$ be a directed, unweighted graph. A topological sorting exists only if the graph is acyclic.
☒ Correct ☐ Incorrect ✓

2.5 / 4

NAME: _____

**Aufgabe 6: Connected Components** (0.5 + 1 + 1 + 1 + 0.5 Punkte)

Consider the following directed Graph.

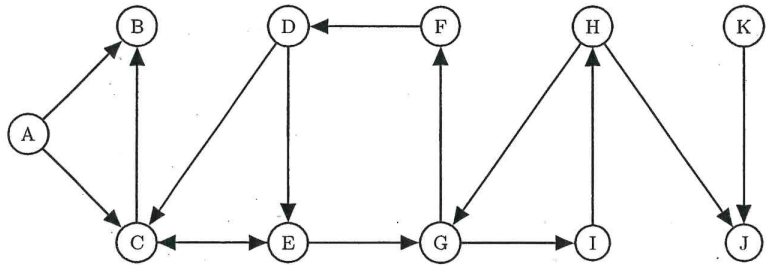

a) How many strongly connected components does this graph have?

4    ✗

b) Draw the corresponding component graph.

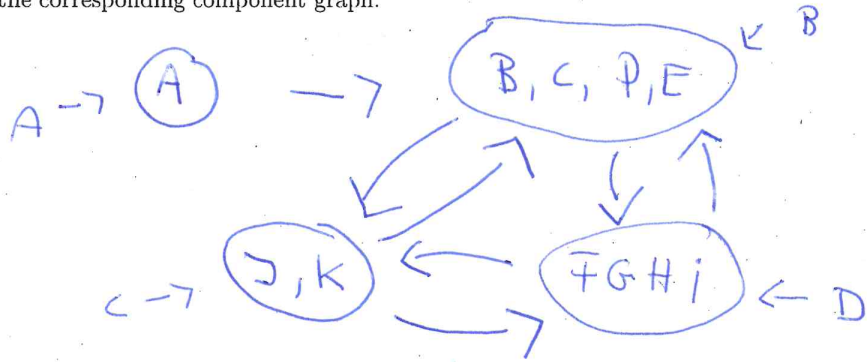

c) Number the different nodes of the above component graph according to the alphabet, i.e., $A$, $B$, $C$, etc. Then specify a topological sort of the component graph.

A C D B    ✗

d) Which nodes of the above component graph are source nodes, which are sink nodes?    Folgefehler

Source nodes: A

Sink nodes: C    0,5

e) In a directed, acyclic graph, the component graph corresponds to the original graph. ✓

☒ Correct         ☐ Incorrect

1/4 Pkt.

NAME: _____

**Aufgabe 7: Shortest Paths** (1+1+0.5+0.5+1 Punkte)

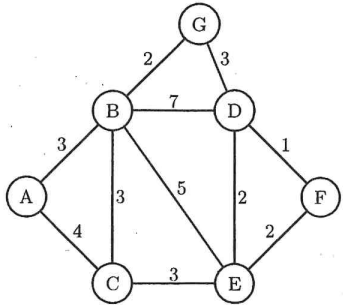

a) Perform the Dijkstra algorithm on the above graph to find the shortest path from A to all other nodes. Give the order in which the Dijkstra algorithm visits the nodes.

A  B  G  C  E  D  F

b) Now give the length of the shortest path from A to all other nodes.

| B | C | D | E | F | G |
|---|---|---|---|---|---|
| 3 | 6/4 | 10/8 | 9/8 | 11/9 | 5 |

c) If a graph does not contain a negative weight cycle, then the result of the Dijkstra algorithm is correct.

☒ Correct   ☐ Incorrect

d) For graphs with $\Theta(|V|^2)$ many edges, the Dijkstra algorithm is best implemented using a

☒ Array.   ☒ Heap.

e) Let $G = (V, E)$ be a directed graph. Now assume we define a weight function on the nodes, i.e., $c : V \to \mathbb{R}_{\geq 0}$. Does it still make sense to speak of the shortest path problem with a weight function on the nodes? If Yes, then give a formal definition of the problem. If No, then explain why the problem makes no sense.

It is not meaningful to speak of the shortest path problem with a weight function on the nodes, as the weight function is typically defined on the edges in the problem. The shortest paths problem seeks to find the shortest path between two nodes in a graph, where the distance between nodes is defined by the sum of the weights of the edges on the path between them.

1st Exam "Algorithms and Data Structures", February 2023     8 / 15

However, if we define a weight function on the nodes, we can modify the shortest path problem to account for this by incorporating the node weights into the edge weights. Specifically, we can define the weight of an edge $(u,v)$ to be the sum of the weight of the starting node $u$ and the weight of the ending node $v$, i.e. $w(u,v) = c(u) + c(v)$. Then, we can solve the modified shortest path problem using any standard shortest path algorithm, such as Dijkstra's algorithm or Bellman-Ford algorithm.

seen

NAME: _____

**Aufgabe 8: Algorithms**  (0.5+0.5+1+0.5+0.5+1 Punkte)

a) What is the Best-Case running time of the following algorithm in relation to the length $n$ of the input array A?

    **function** ARRAYFUNCTION(Array A)  
        **for** i=1,...,A.length **do**  
            A[i] = i  
        **end for**  
        A = QuickSortWithLastElementAsPivot(A)  
        **return** A  
    **end function**

Best-Case: $\Theta(n \log n)$    f

b) The following program calculates a hash function. Which hash function is calculated by the program?

    **function** HASHFUNCTION(Integer $n$)  
        $n = n + 2$  
        $n = 7 * n$  
        **while** $n \geq 153$ **do**  
            $n = n - 153$  
        **end while**  
        **return** n  
    **end function**

Hashfunktion: $7 \cdot (n+2) \bmod 153$  ✓

c) What are the Best-Case and Worst-Case running time of the following algorithm in relation to the length $n$ of the input array A?

    **function** ARRAYFUNCTION(Array A)  
        **for** $j = 2$ to A.length **do**  
            key = A[j]  
            $i = j - 1$  
            **while** $i > 0$ and A[i]>key **do**  
                $A[i+1] = A[i]$  
                $i = i - 1$  
            **end while**  
            $A[i+1] = $ key  
        **end for**  
        **return** A  
    **end function**

Best-Case: $\Theta(n)$

Worst-Case: $\Theta(n^2)$  ✓

d) What is a Worst-Case input sequence of length $n$ for the algorithm in c)?

$n, n-1, \ldots, 3, 2, 1$  ✓

NAME: _____

(Continuation of Exercise)

e) What is the Worst-Case running time of the following algorithm in relation to the length $n$ of the input array A?

    **function** MYALGORITHM(Array $A$)
        **if** A.length $\leq$ 100 **then**
            **return** $A[1]$
        **end if**
        Result = $4 \cdot$ MyAlgorithm($A[1\ ...\ $A.length$-1]$)
        **return** Result
    **end function**

Worst-Case: $\Theta(\ n\ )$ ✓

Hint: $A[1\ ...\ $A.length$-1]$ is the Array A without the last element. This operation takes time $O(1)$.

f) The diameter of an undirected and unweighted graph corresponds to the largest distance between two nodes in a graph. More specifically, the diameter is

$$\text{diam}(G) = \max_{(u,v) \in V \times V} \{\text{dist}(u,v)\},$$

where $\text{dist}(u,v)$ corresponds to the length of the shortest path between $u$ and $v$. Provide an algorithm that calculates the diameter of a graph $G = (V, E)$. Describe your algorithm in pseudo-code.

```
Function diameter (graph G):
    max-distance = 0
    for each vertex v in G:
        distances = BFS(G, v)    // or DFS(G,v)   -0.5
        max-distance = max(max-distance,
                           max(distances))
    return max-distance
```

This Algorithm iterates through all vertices in the Graph and performs BFS to find the shortest path to all other vertices. It then updates the maximum distance found so far, and returns it as the diameter of the graph.

NAME:

**Aufgabe 9: Matrix Travel** (2+2 Punkte)

Given an MxN matrix A of positive numbers, we can imagine this matrix as a game board of size MxN. The entry at $(i, j)$ corresponds to the costs that must be paid to enter field $(i, j)$ of the game board. We now want to move a game piece as cost-effectively as possible from the upper left field $(1, 1)$ to the lower right field $(N, M)$. In each step, the game piece may only move one field down or one field to the right. So it can go from field $(i, j)$ to field $(i + 1, j)$ or to field $(i, j + 1)$. We are interested in the costs of the cheapest path from $(1, 1)$ to $(N, M)$. In the following example, the cheapest path costs 33.

$$A = \begin{pmatrix} 4 & 7 & 8 & 4 & 6 \\ 6 & 7 & 3 & 9 & 2 \\ 3 & 8 & 1 & 2 & 4 \\ 7 & 1 & 7 & 3 & 7 \\ 2 & 9 & 8 & 9 & 3 \end{pmatrix}, \quad \text{A cheapest path } A = \begin{pmatrix} 1 & 7 & 8 & 4 & 6 \\ 6-7-3 & 9 & 2 \\ 3 & 8 & 1-2 & 4 \\ 7 & 1 & 7 & 3-7 \\ 2 & 9 & 8 & 9 & 3 \end{pmatrix}$$

a) This problem can be solved with dynamic programming. Set up an appropriate recursive equation.

T is minimum cost to reach the cell

$$T[1,1] = A[1,1] \quad , \quad T[i,1] = \sum_{k=1}^{i} A[k,1] \quad , \quad T[1,j] = \sum_{k=1}^{j} A[1,k]$$

For i, j > 1:

$$T[i,j] = A[i,j] + \min\left(T[i-1,j], T[i,j-1]\right) \quad \checkmark$$

b) Complete the following pseudo-code to solve the problem in runtime $O(MN)$.

```
function MATRIXTRAVEL(Cost matrix A)
    T = MxN matrix filled with zeros
    for i = 1 to N do
        for j = 1 to M do
            if i=1 and j=1:
                T(1,1) = A(1,1)
            elsif i=1:
                T(1,j) = A(1,j) + T(1,j-1)
            elsif j=1:
                T(i,1) = A(i,1) + T(i-1,1)
            else:
                T(i,j) = A(i,j) + min(T(i-1,j), T(i,j-1))
            endif
        end for
    end for
    return T[N,M]
end function
```
✓

NAME: _____

**Aufgabe 10:  Prove or Disprove**  (4 Punkte)

a) Let $G = (V, E)$ be a connected, undirected and weighted graph. A cut $(S, V \setminus S)$ is a partition of the nodes (that is, we divide the nodes into two groups, not necessarily of equal size). A cut edge is an edge $(u, v)$ with one node in $S$ and one node in $V \setminus S$, that is $u \in S$ and $v \in V \setminus S$. Prove or disprove the following statement:

*Assume that $G$ has a uniquely determined minimum spanning tree. Then, for each cut $(S, V \setminus S)$ of $G$, there exists a uniquely determined cut edge with minimum weight.*

The statement is true.

<u>Proof</u>: Let $T$ be the unique MST of $G$. Consider any cut $(S, V\setminus S)$ of $G$. We want to prove that there exists a unique edge with minimum weight that crosses this cut.

Assume (for contradiction) that there are two different cut edges $(u,v)$ and $(x,y)$ with the same minimum weight. W.l.o.g., assume $u, x \in S$. There exists a path in $T$ between $u$ and $x$. Let $z$ be the first node on this path that is in $V\setminus S$. Then, the edge $(z,x)$ must be in the unique path in $T$ between $u$ and $x$. Therefore, the weight of $(z,x)$ is less or equal to the weight of $(u,v)$ which is the same as the weight of $(x,y)$. This contradicts the assumption that $(u,v)$ and $(x,y)$ have the same minimum weight.

Therefore, there can be at most one cut edge with minimum weight that crosses the cut $(S, V\setminus S)$. To show that there exists such an edge, we can consider the unique path in $T$ between any node in $S$ and any node in $V\setminus S$. This path must contain at least one edge that crosses the cut $(S, V\setminus S)$.

↱ next page

and the edge with minimum weight among these edges is the unique cut edge with minimum weight that crosses the cut.

Therefore, for each cut $(S, V\setminus S)$ of $G$, there exist a uniquely determined cut edge, as required.

1/4 Pkt.



\end{document}